\documentclass[letter, 10 pt, journal, twoside]{IEEEtran}

\IEEEoverridecommandlockouts                              

\usepackage[utf8]{inputenc}
\usepackage[T1]{fontenc}

\title{ \LARGE \textbf{Developing cooperative policies for multi-stage reinforcement learning tasks}}

\author{Jordan Erskine$^{1}$, Chris Lehnert$^{2}$%
\thanks{Manuscript received: December, 23, 2021; Revised March, 26, 2022; Accepted April, 19, 2022.}
\thanks{This paper was recommended for publication by Editor Jens Kober upon evaluation of the Associate Editor and Reviewers’ comments. [Note that the Editor is the Senior Editor who communicated the decision; this is not necessarily the same as the Editor-in-Chief.]}
\thanks{$^{1}$J. Erskine and C. Lehnert are with the Queensland University of Technology (QUT), Brisbane, Australia and affiliated with Queensland Centre of Robotics (QCR) {\tt\small jordan.erskine@hdr.qut.edu.au, c.lehnert@qut.edu.au}}
\thanks{Digital Object Identifier (DOI): see top of this page.
}
}

\usepackage{comment}
\usepackage{cite}
\usepackage{amsmath}
\usepackage{amssymb}
\usepackage{graphicx}
\usepackage{booktabs}
\usepackage{makecell}
\usepackage{float}
\usepackage[ruled, noline]{algorithm2e}
\usepackage{multirow}
\usepackage{xcolor}

\usepackage{lipsum}

\begin{document}
\maketitle


\begin{abstract}
    \textbf{Many hierarchical reinforcement learning algorithms utilise a series of independent skills as a basis to solve tasks at a higher level of reasoning. These algorithms don't consider the value of using skills that are cooperative instead of independent. This paper proposes the Cooperative Consecutive Policies (CCP) method of enabling consecutive agents to cooperatively solve long time horizon multi-stage tasks. This method is achieved by modifying the policy of each agent to maximise both the current and next agent's critic. Cooperatively maximising critics allows each agent to take actions that are beneficial for its task as well as subsequent tasks. Using this method in a multi-room maze domain and a peg in hole manipulation domain, the cooperative policies were able to outperform a set of naive policies, a single agent trained across the entire domain, as well as another sequential HRL algorithm.}
\end{abstract}


\section{Introduction}


\IEEEPARstart{M}{any} of the struggles that Reinforcement Learning (RL) faces have been addressed within the field of Heirarchical Reinforcement Learning (HRL) \cite{nachum2019does}. Complex domains such as ant \cite{nachum2018near}, humanoid \cite{peng2019mcp} and swimmer \cite{li2019hierarchical} have all had high levels of success with hierarchical methods while non-hierarchical methods have struggled to make progress. Hierarchical methods take advantage of the ability of many tasks to be abstracted, solving the problem of high level reasoning separately from low level control \cite{wulfmeier2021data}. 

Many HRL approaches involve developing a series of skills (low level policies) that are used by a high-level controller to solve tasks \cite{vezhnevets2017feudal}.  However, these skills cannot always be utilised by a high level controller to optimally solve a task \cite{nachum2018data}. For example, for the task of cutting a cake, a high level controller may employ the skill of grasping a knife, followed by manoeuvring the knife above the cake, and finally lowering the knife to cut the cake. A skill trained to grasp a knife may choose to grasp the knife by the blade, a behaviour appropriate for other tasks, such as passing a knife safely to a human. This method of grasping however would not allow for effective cake cutting. The solution to this problem that is proposed in this paper is to use policies that are aware of the next policy's goal.

We propose a method of training agents to accommodate for consecutive agents. This method allows a collection of agents to work together more cohesively to complete multi-stage tasks. In our proposed approach, each agent is incentivised to cooperate with the next agent by training the agent's policy network to produce actions that maximise both the current agent's critic and the next agent's critic, weighted by an introduced parameter, the cooperative ratio. By incorporating the next agent's critic, the current agent can continue to achieve its own goal while also producing a solution that is beneficial for the next agent.

The contributions produces by this paper are:
\begin{itemize}
    \item A novel method, Cooperative Consecutive Policies (CCP), which enables agents to learn behaviours that maximise reward for their own task while accommodating for subsequent tasks, improving performance for learning multi-stage tasks.
    \item Two case studies in a continuous state/action space maze domain and a robotic manipulation domain. In these experiments, CCP outperformed a set of naive policies (trained to greedily maximise subtask reward), a single agent trained end-to-end on the same task, as well as sequential HRL baseline, using the transition policies method. 
    \item An ablation study on the effect of varying the cooperative ratio. This includes a study on the effect of the cooperative ratio on success rate, as well as the ability to use different cooperative ratios for different cooperative policies.
\end{itemize}

\begin{figure}
    \centering
    \includegraphics[width = \linewidth]{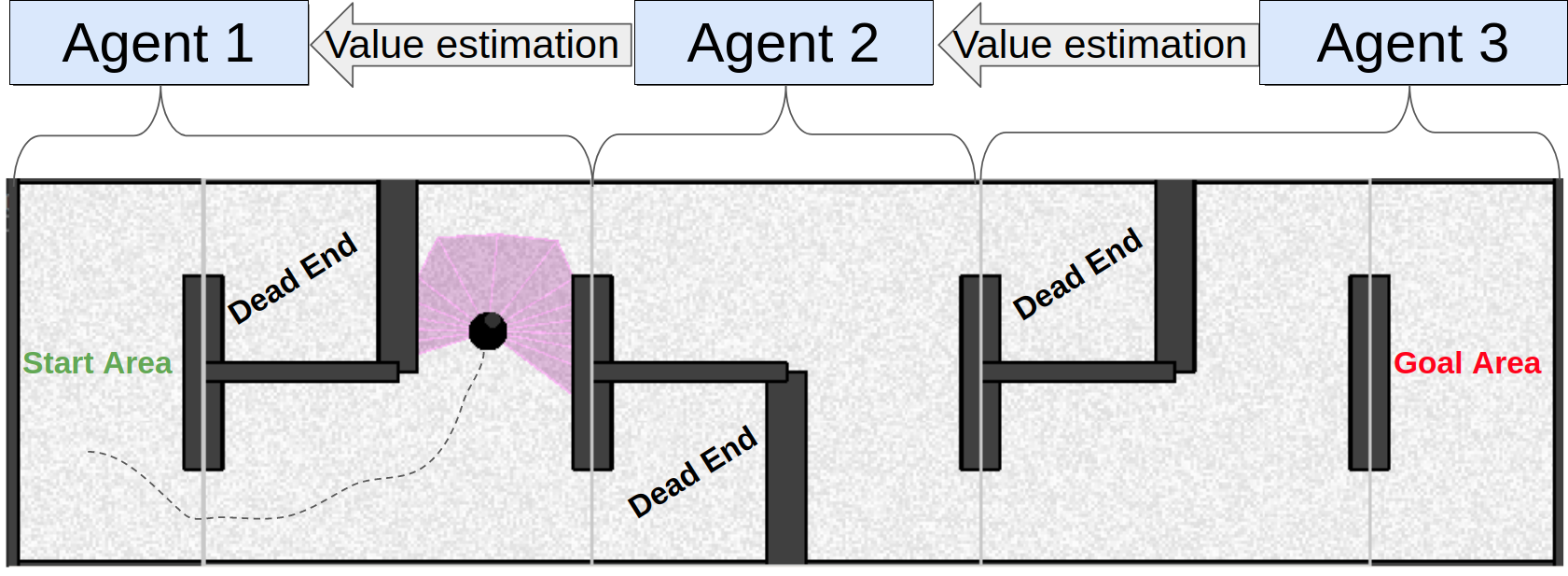}
    \caption{A high level view of the Cooperative Consecutive Policies method works. Each agent acts within its own subtask, and is informed by the next agent about how to act to assist in completing the subsequent task. In the maze domain, the next agent can give information about where the dead ends are and how to avoid them. }
    \label{fig:overview}
\end{figure}

\begin{figure*} 
    \centering
    \includegraphics[width=\linewidth]{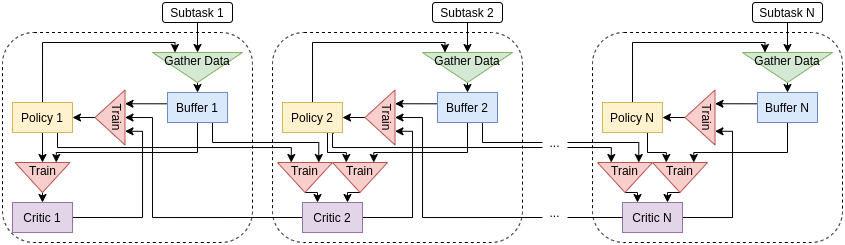}
    \caption{The structure of a cooperative policy implementation in a N subtask environment. Each policy is trained using it's own critic and the subsequent agent's critic, though the last agent is only trained relative to its own critic. Each critic is trained using data from the current and previous subtasks, utilising the policy from the agent corresponding to that subtask.}
    \label{fig:CoopPolDiagram}
\end{figure*}

\section{Related Work}

Hierarchical Reinforcement Learning (HRL) has been a long standing field in RL \cite{al2015hierarchical}. HRL methods capitalise on the inherent structure that is present in many tasks. An important benefit of using HRL is that learning a series of smaller, simpler skills is easier and faster to learn than learning to solve a single, more complex task \cite{sutton2018reinforcement} \cite{frans2017meta} \cite{iqbal2019actor} .

Some hierarchical reinforcement learning approaches have been designed to deal with sequential tasks \cite{oh2017zero} \cite{icarte2018using}. Typically they have the form of using a meta-controller in conjunction with a series of lower-level policies for each subtask \cite{bacon2017option} \cite{azarafrooz2019hierarchical} \cite{gupta2019relay} \cite{tessler2017deep}, which allows for more generalisation, as the series of subtasks can be combined in more versatile ways. These methods follow one of two approaches; manual definition of each subtask, or learning of the subtasks. 

Many HRL algorithms seek to learn skills autonomously. These skills are simple behaviours that can be utilised by the high level meta-controller \cite{Andreas2017} \cite{haarnoja2018latent} \cite{esteban2019hierarchical} \cite{sharma2019dynamics}. This approach ensures that the skills are useful for the overall task, but this tends to be very slow to train. 

Manually defining subtasks is an another approach to HRL. This approach involves decomposing a task into subtasks manually, defining the reward signal and termination signal for each subtask \cite{Andreas2017} \cite{forestier2017intrinsically}. Separate agents are then trained to solve these subtasks independently. The fact that these agents are not incentivised to assist in solving the overall task means that without careful engineering of subtasks, suboptimal solutions to tasks are likely \cite{florensa2017stochastic}. Our method is designed to overcome the difficulties of engineering subtasks by enabling the agents to cooperate towards completing the task, loosening the requirements of careful design. There are other methods that seek to do a solve a similar problem by learning transition policies between previously learnt subtasks \cite{lee2018composing} \cite{lee2021adversarial}, but these transition policies may still struggle if the solution to one subtask is too far from an adequate starting position for the subsequent subtask.

\section{Method}

This paper introduces the Cooperative Consecutive Policies (CCP) method. In the CCP method, a cooperative policy takes into account the subsequent agent's critic. The critic informs the policy of how good certain actions and states are to complete its task effectively. Using both the subsequent and current critic allows a policy to act to solve the current subtask in a way that allows an effective solution in future subtasks.

The CCP method is not an algorithm in and of itself. It is an algorithmic change that can be applied to any actor critic algorithm, assuming there are consecutive agents used to solve a task. The experiments described in this paper were implemented using the SAC algorithm \cite{haarnoja2018soft} \cite{haarnoja2018soft2}.

\subsection{Procedure}
We consider a modified MDP formulation for solving the problem of picking optimal actions to solve a task. We assume an environment that involves a task that is decomposed into a series of \(N\) subtasks. At each timestep \(t\) an agent can take an action \(a_t\) from the current state \(s_t\), which results in the environment evolving to the next state \(s_{t+1}\), producing a transition signal \(U(s_t) \in [1,N]\) that determines what subtask is currently active, and produces a series of \(N\) reward signals \(r_{n,t} |n \in [1,N]\) that correspond to each subtask.

The CCP method requires creating \(N\) agents, one for each subtask, that each consist of a policy \(\pi_n|n \in [1,N]\) with parameters \(\theta_n\) and a critic \(Q_n|n \in [1,N]\) with parameters \(\beta_n\). This method requires that the subsequent agent is known to the current agent. Each critic maximises the discounted sum of future rewards \(r_n\) from its subtask. Each policy, rather than maximising the critic that corresponds to their own subtask, instead maximises a convex sum of the current and subsequent critics. The standard critic loss signal (Equation \ref{eq:criticUpdate}) and the updated policy loss signal (Equation \ref{eq:policyUpdate}) are,
\begin{equation} \label{eq:criticUpdate}
    \nabla_{\beta_n}=\frac{1}{M}\sum_0^M(Q_n(s,a) - (r_n + \gamma Q_n(s',\pi_n(s')))^2
\end{equation}


\begin{equation} \label{eq:policyUpdate}
    \nabla_{\theta_{n}} = \frac{1}{M}\sum^{M}_{0}( - C(Q_n,Q_{n+1}))
\end{equation}

\color{black}
Where \(M\) is the number of samples in a batch \(b\) sampled from replay buffer \(B_n\), \(\alpha\) is the entropy maximisation term, and \(C\) is a convex combination of the current and subsequent critics, as defined by:
\begin{equation} \label{eq:convComb}
    C(Q_n,Q_{n+1}) = \eta \hat Q_n(s,\pi_n(s))+(1-\eta) \hat Q_{n+1}(s,\pi_n(s))
\end{equation}
Where \(\eta\) is the cooperative ratio. This ratio affects how much the current policy acts with respect to the subsequent critic, and is a number between 0 and 1. A cooperative ratio closer to 1 is incentivised to maximise the current critic's estimate, whereas a cooperative ratio closer to 0 is incentivised to maximise the subsequent critic's estimate. Each Q function is normalised across the batch as denoted by \(\hat Q\) using:
\begin{equation} \label{eq:normalise}
    \hat Q(s,a) = \frac{Q(s,a)-\displaystyle\min_{s'\in b}Q(s',\pi(s'))}{\displaystyle\max_{s'\in b}Q(s',\pi(s'))-\displaystyle\min_{s'\in b}Q(s',\pi(s'))}
\end{equation}
This normalisation is done for each critic separately and is recalculated for each batch. It bounds the critics output to values between 0 and 1. This is required to compare the current and subsequent critics, which can produce diverse ranges of values during training. The full algorithmic approach involves simultaneously gathering data (as shown in Algorithm \ref{alg:gather}) and training on that data (as shown in Algorithm \ref{alg:train}). 
Refer to appendix (Section \ref{app:proof}) for mathematical analysis of this method.


\color{black}
This implementation of CCP is designed using SAC as the base algorithm. To apply CCP to other algorithms is done by modifying the algorithm's policy update to using \(C\) in place of the traditional critic evaluation, and ensuring each agent is updated using the correct buffer \(B\), as shown in Figure \ref{fig:CoopPolDiagram}. 


\begin{algorithm} [t]
\caption{Gathering Data}
\label{alg:gather}
\SetAlgoLined
    Environment with N subtasks and associated reward signals \(r_{(1, ..., N)}\)\;
    For each subtask initialise an agent \(A_n\), including a policy \(\pi_n\) with parameters \(\theta_n\), a critic \(Q_n\) with parameters \(\beta_n\), and a replay buffer \(B_n\)\;
    \While{timestep \(<\) maxTimestep}{
        \(s, n \xleftarrow[]{}\) reset environment\;
        \While{not done}{
            \(a\) \(\sim \pi_n(s)\)\;
            \(s', r_{(1,...,N)}, done \xleftarrow[]{}\) environment step with \(a\)\;
            record \((s, a, r_{1,...,N}, s', done)\) in \(B_n\)\;
            \(s \xleftarrow[]{} s'\)\;
            \(n \xleftarrow[]{} U(s')\)\;
        }
        
    }

\end{algorithm}

\begin{algorithm}
\caption{Cooperative Training}
\label{alg:train}
\SetAlgoLined
    Set of N agents \(A_n\), each with policy \(\pi_n\) with parameters \(\theta_n\), a critic \(Q_n\) with parameters \(\beta_n\), and a replay buffer \(B_n\)\;
    cooperative ratio \(\eta\)\;
    discount factor \(\gamma\)\;
    entropy maximisation term \(\alpha\)\;
    \For{n in (1,...,N)}{
        sample minibatch \(b\) from \(B_n\) of M samples \(\xrightarrow{}(s, a, r_{(1,...,N)}, s', d)\)\;
        \For{j in (n, n+1)}{
            \(y(r_j,s',d) = r_j + (1-d)\gamma(Q_j(s',a') - \alpha\) log(\(\pi_n(a'|s')))\), \(a'\sim \pi_n(s')\)\;
            \(\nabla_{\beta_j}=\frac{1}{M}\sum(Q_j(s,a) - y(r_j,s',d))^2\)\;
        }
        \(a' \sim \pi_n(s)\)\;
        \For{j in (n, n+1)}{
            \textit{Calculate \(\hat Q_n\) across minibatch \(b\) according to Equation \ref{eq:normalise}}\;
        }
        \(C(Q_n,Q_{n+1}) = \eta \hat Q_n(s,a') + (1-\eta)\hat Q_{n+1}(s,a')\)\;
        
        \(\nabla_{\theta_{n}}\) = \(\frac{1}{M}\sum(\alpha \log\pi_n(a'|s) - C(Q_n,Q_{n+1}))\)\;
        
    }
\end{algorithm}

\section{Experimental studies}


\begin{figure}[t]
    \centering
    \includegraphics[width = \linewidth]{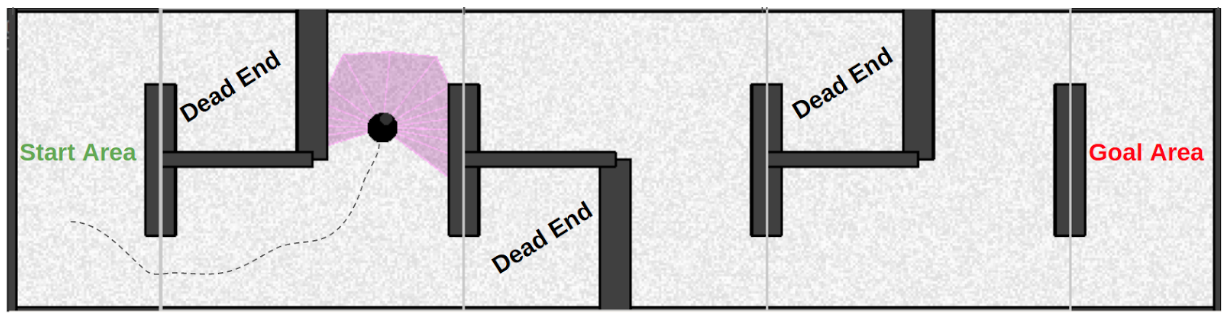}
    \caption{A 3 room example of the Maze domain. The first agent begins in the starting area and produces actions to navigate the first room. As the agent enters the next room, the next takes charge to navigate that room. The agent's scan range is shown in purple.}
    \label{fig:MazeDiag}
\end{figure}

\subsection{Environments}
Two domains were used to test the efficacy of the CCP method: the maze domain, and the peg in hole domain.

\subsubsection{Maze Navigation}
In the maze domain, a series of consecutive rooms were created. Each room had two paths to exit into the next room, one of which leads to a dead end. Each room was considered a subtask with the goal being to exit the room, with a reward signal dependent on horizontal position, starting at 0 on the left of the room and linearly increasing to 1 at the right of the room. The overall task in this domain was to travel through all the rooms to get from one side to the other. The rooms are designed such that an optimal subtask solution would not lead to an optimal overall solution, as the reward signal for each task increases as the position moves to the right, not towards the correct path. This domain allows for easy extension in terms of the number of subtasks that are required to be able to solve the overall task. 


The agent in this domain uses two continuous actions: linear velocity and angular velocity. The agent uses an observation of the environment that includes a laser scan as well as a global position in the maze, all of which are continuous measurements. This domain is shown in Figure \ref{fig:MazeDiag}. 

\subsubsection{Peg in hole manipulation}

The second domain used was the peg in hole domain 
. In this domain, the task is to use a Panda arm to insert a peg into a hole. This domain is decomposed into two subtasks: grasping and inserting. The grasping subtask requires the gripper to establish a grasp on the peg. This subtask uses an exponentially decaying reward based on distance between the centre of the peg and the fingers of the gripper. The second subtask is to move the peg towards and then into the hole. The reward for this subtask is also an exponentially decaying reward, this time based on the distance between the centre of the peg and the centre of the hole. The task is considered a success if the peg is fully inserted into the hole.

This domain was chosen as it represents a real world example of how completing one subtask can affect the completion of a second subtask. If the peg is grasped by the thin section, then the insertion subtask cannot be completed optimally, as the peg cannot be fully inserted into the hole. This domain is more complex than the maze domain, due to the higher dimensionality of the state space and the complexity of grasping dynamics

The agent in this domain has two continuous actions: end effector velocity along the axis of actuation and gripper force. The state that the agent observes is the positions, orientations and velocities of the peg, hole and end effector. 

Both domains in this paper use relatively simple reward signals. Though more complex, and potentially more informative, reward signals could be constructed, these simple reward signals were used on purpose to show that using a imperfect reward signal can be overcome using the CCP method. Engineering a reward signal that can effectively avoid globally suboptimal behaviours is an expensive and sometimes intractable problem, and being able to solve a task without it is a valuable quality.


\subsection{Algorithms}

Four different methods were evaluated in these domains:
\begin{itemize}
    \item \textbf{CSAC:} The CSAC method is an implementation of Soft Actor Critic (SAC) \cite{haarnoja2018soft}\cite{haarnoja2018soft2} that utilises the CCP methodology of incentivising cooperative solving of sequential tasks, making a set of Cooperative SAC agents.
    \item \textbf{Naive SAC:} The naive method uses a separate SAC agent for each subtask, without any communication between agents. Each agent is attempting to maximise solely its own reward signal. This method represents the naive approach of treating each agent independently.
    \item \textbf{SAC:} The SAC method involves using a single end-to-end SAC agent trained to perform across the whole domain, utilising a reward signal that is the combined reward signal from all the subtasks. This method represents the standard RL approach to solving a task. 
    \item \textbf{TP:} The Transition Policies method \cite{lee2018composing} used as a baseline. This method trains primitive policies to complete each subtask, and then trains transition policies to move from subtask termination states to states that are good initialisations for the subsequent subtask.
\end{itemize}
  
Hyperparameters and further implementation details are listed in Appendix 2 (Section \ref{app:hyper}).

\subsection{Studies}
\subsubsection{Optimal cooperative ratio}

The first study tested the performance of the four different methods in both the maze domain and the peg in hole domain. The CCP method requires the selection of a cooperative ratio to determine the trade-off between current and subsequent critics. In the maze domain three different length mazes were tested using a sweep across the cooperative ratio parameter to determine its effects on learning performance, with a similar sweep conducted in the peg in hole domain. The best results from these sweeps were then compared to the results from using the other methods. This study also investigated the sensitivity of the method's performance with respect to the cooperative ratio.

\subsubsection{Independent cooperative ratios}

The previous experiment used the same cooperative ratio for each agent within each experiment. An investigation was conducted to determine whether using independent cooperative ratios for each agent would improve the performance, conducted in the 3 room maze domain, in which there are two cooperative agents. A parameter sweep was conducted across the cooperative ratio for each cooperative agent to investigate it's effect on learning performance.

\section{Results}

\begin{figure*}
    \centering
    \includegraphics[width=\linewidth]{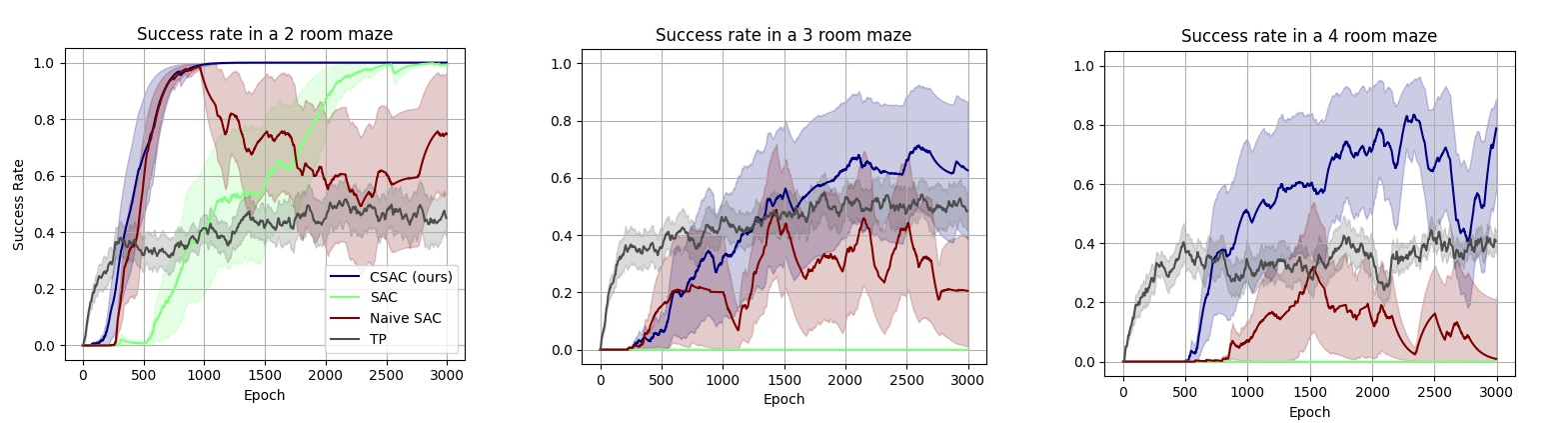}
    \caption{Success rate in the 2, 3 and 4 room maze domains. Each plot shows the average success rate across several different random seeds, where success is defined as reaching the end of the maze. The domains use a cooperative ratio of 0.1, 0.1 and 0.9 respectively.}
    \label{fig:MazeResults}
\end{figure*}

\begin{figure}
    \centering
    \includegraphics[width=0.75\linewidth]{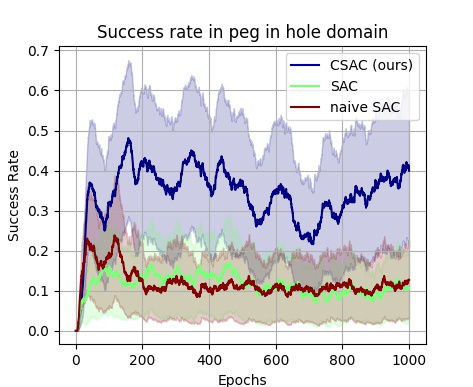}
    \caption{Success rate in the peg in hole domain. The plot shows the average success rate across several different random seeds. This uses a cooperative ratio of 0.9}
    \label{fig:pegInHoleResults}
\end{figure}

\begin{table}[]
    \begin{center}
    \caption{Success rate of each method in maze and peg in hole domains}
    \label{tab:results}
    \begin{tabular}{cccccc}
        \toprule
        \multirow{2}{*}{\textbf{Method}} & \multirow{2}{*}{\textbf{\makecell{Cooperative \\Ratio}}} & \multicolumn{3}{c}{\textbf{Maze - Subtasks}} & \multirow{2}{*}{\textbf{Peg in hole}} \\
        & & 2 & 3 & 4 & \\
        \midrule
        \multirow{9}{*}{CSAC} & 0.1 & \textbf{1.00} & \textbf{0.62} & 0.00 & 0.11 \\
        & 0.2 & \textbf{1.00} & 0.23 & 0.00 & 0.11 \\
        & 0.3 & \textbf{1.00} & 0.07 & 0.00  & 0.11 \\
        & 0.4 & 0.31 & 0.00 & 0.00 & 0.17 \\
        & 0.5 & 0.05 & 0.06 & 0.00 & 0.21 \\
        & 0.6 & 0.17 & 0.04 & 0.00 & 0.34 \\
        & 0.7 & 0.42 & 0.40 & 0.17 & 0.31 \\
        & 0.8 & 0.51 & 0.16 & 0.50 & 0.37 \\
        & 0.9 & 0.06 & 0.21 & \textbf{0.79} & \textbf{0.40}\\
        \midrule
        \multicolumn{2}{c}{SAC} & \textbf{1.00} & 0.00 & 0.00 & 0.11\\
        \midrule
        \multicolumn{2}{c}{Naive} & 0.54 & 0.20 & 0.14 & 0.13\\
        \midrule
        \multicolumn{2}{c}{TP} & 0.43 & 0.50 & 0.40 & - \\
        \bottomrule
    
    \end{tabular}
    \end{center}
\end{table}

\subsection{Optimal cooperative ratio}
\subsubsection{Maze navigation}
The results for the 2 room experiment within the maze domain are presented in Figure \ref{fig:MazeResults} and are summarised in Table \ref{tab:results}. This experiment shows that the cooperative and naive agents both learn a successful policy in a similar time period, whereas the single agent policy took more than 3 times as long to reach a similar performance. The naive agents, though they reached a high level of success quickly, had a decaying performance. This is due to the fact that each agent learned a solution to their domain quickly, which included travelling to the further door. As each agent refined its solution, the shorter path, which is suboptimal overall, was used more frequently, reducing the performance of the overall task. This experiment shows that decomposing a task into subtasks is beneficial in terms of training speed, shown by the relative training speed of the cooperative and naive policies compared to the single agent. This experiment also shows that just decomposing a task into subtasks and then treating them as entirely separate problems can lead to suboptimal or decaying solutions. 

Figures \ref{fig:MazeResults} and Table \ref{tab:results} show the success rates of the three different agent types in the 3 and 4 room mazes, where success is defined as reaching the end of the maze. In both of these domains, the single agent was not able to learn a successful policy at all within the training window. This shows that the local minima, the dead end paths, proved too difficult for the end-to-end SAC agent to overcome within the time frame. The cooperative agents were able to achieve a consistently higher performance than the naive agents. This demonstrates the value of splitting a task into subtasks and then solving them cooperatively. 

Across all room configurations the TP baseline had a suboptimal level of performance. This method learns a set of transition policies that attempts to manipulate the agent from the termination state of a subtask to a good starting state for the subsequent subtask. Due to the way the doorways are arranged within the maze, a subtask that concludes at a dead end is too far from a good starting state for the next subtask for a transition policy to be able to rectify.


\subsubsection{Peg in hole manipulation}

Figure \ref{fig:pegInHoleResults} shows the results of using each of the 3 methods in the peg in hole domain. CSAC is able reach a far higher level of performance in this domain compared to both the single agent SAC method and the naive method. The low success rate in this domain is due to the difficulty of using contact physics. If the agent grasps the peg in the wrong way or applies too much force, the peg can be pushed into a pose that is unreachable for the robot, making the episode unsolvable. Despite this, CSAC is able to complete the task 40\% of the time, far greater than the other two methods which only achieved a success rate of 11\% and 13\% for single SAC and naive SAC respectively. 

The SAC method has a low performance in this domain due to the fact that it is encouraged to move the peg to the hole as quickly as possible. This requires that the peg be grasped by the narrow section and dragged to the hole. Due to this poor grasp location, the task cannot be fully solved. In the case of the naive SAC, the reward signal that the first agent is getting is to move close to the peg, not necessarily to grasp it. This means that the first agent will hover close to the peg to maximise it's reward. The CCP method avoids both these behavioural pitfalls and completes the task with a relatively high level of success.


Table \ref{tab:results} shows the different success rates in each domain using a range of cooperative ratios. These results show that the cooperative ratio is a sensitive variable, and it's optimal value is task dependent. This shows that is important to use a sweep across the variable to ensure a good performance. 

\subsection{Independent cooperative ratios}

\begin{figure}
    \centering
    \includegraphics[width = 0.8\linewidth]{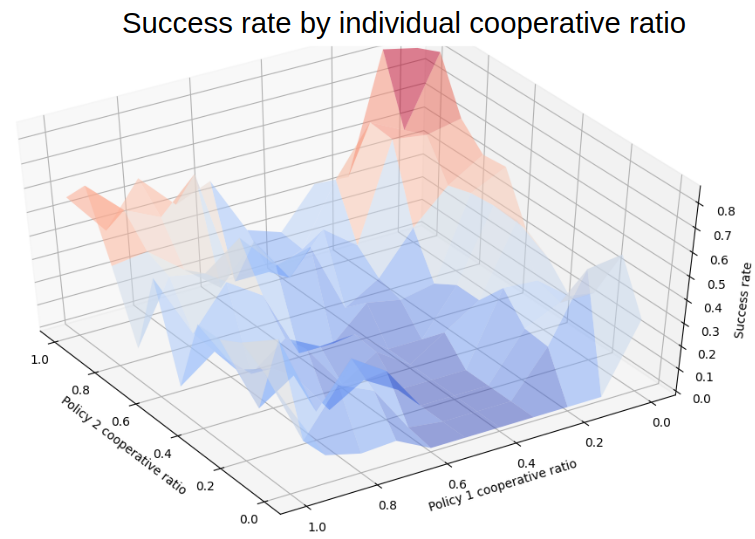}
    \caption{The measured success rate in the 3 room maze domain with different cooperative ratios for each cooperative policy. The success rate is an average success rate across the last 10 epochs of each of the 10 different randomly initialised iterations of each configuration. Blue represents a low success rate and red represents a high success rate.}
    \label{fig:indCRatios}
\end{figure}

The results of the individual cooperative ratio experiments are shown in Figure \ref{fig:indCRatios}. This figure shows the average success rate of a set of cooperative policies in a 3 room domain. In this experiment the first and second policies used independent cooperative ratios, represented along the bottom two axes of Figure \ref{fig:indCRatios}. Using the individually tuned cooperative ratios, a higher level of performance was found in the 3 room domain compared to using the same cooperative ratio for both policies. Using cooperative ratios of 0.1 and 1.0 for the first two policies respectively result in the highest level of success (78\%), outperforming the policies trained using a shared cooperative ratio (54\%).

Some insights can be gathered from analysing the effect on success rate when changing the cooperative ratio by looking at Table \ref{tab:results} and Figure \ref{fig:indCRatios}. In all of the shared ratio experiments, using a cooperative ratio around 0.5 produces a low level of performance. In the maze domain, the optimal behaviour for one room compared to the next room involves utilising different paths. Trying to achieve both objectives equally appears to lead to a poor performance overall, as the policy can't decide which objective to prioritise. This trend can be seen in the individual cooperative ratio experiments when comparing the cooperative ratios for the first policy. Using a cooperative ratio of 0.5 for either policy leads to lower performance as seen in the shared ratio experiments. 

The independent ratio experiments also show that using a low cooperative ratio for the first policy has a higher success rate than using a high cooperative ratio, similar to results in the 2 room results in the shared ratio experiments. This likely shows that in the first subtask it is harder to avoid the dead-end without increased direction from the subsequent critic. 

\section{Conclusion}

This paper introduces the CCP method for cooperatively solving multi-stage tasks. This method was tested using the SAC algorithm (implementation called CSAC) in two different domains, the maze domain and the peg in hole domain, and was compared against three other methods; a SAC agent trained end-to-end across the whole domain, a set of naive agents trained to solve each subtask greedily, and a baseline HRL algorithm for sequential tasks, the Transition Policies algorithm.

The CCP method outperformed each other method in the maze domain, as summarised in Table \ref{tab:results} and shown in Figure \ref{fig:MazeResults}. In the simplest domain (2 room maze), CSAC converged on a solution 4 times faster than the single agent and was able to maintain a high level of performance that the naive policies were not able to maintain, while the TP baseline was unable to solve the domain. In the more complex domains (3 and 4 room mazes), the cooperative policies had a consistently higher level of performance than the naive policies and TP baseline, whereas the single agent was not able to find any solution to the task within 3 million training steps. Similar results were found in the peg in hole domain (Figure \ref{fig:pegInHoleResults}), in which the algorithm using CSAC had a success rate approximately 30\% higher than the other methods.

Currently CCP requires that the task be decomposed into subtasks before training. There are works that seek to learn the decomposition of a task \cite{Andreas2017}, and these methods could be combined with CCP for a more generalisable algorithm. Additionally, the cooperative ratio variable is required to be defined for each subtask. A proposed solution to this is to utilise meta-learning \cite{frans2017meta} to tune this parameter. This method also only provides direction based on the subsequent subtask's critic. If a specific action needs to be taken to solve the overall task and the subsequent agent isn't aware of it, then CCP will still fail. Future works will seek to address these issues and make this method more robust.

\bibliographystyle{IEEEtran}
\bibliography{references.bib}


\section{Appendix 1} \label{app:hyper}

The implementation used in this paper was based on the RLkit implementation of SAC \cite{vitchyr}. This implementation utilises an epoch based approach. All methods use the hyperparameters shown in Table \ref{tab:Hypers}. These hyperparameters were adapted from the RLkit SAC implementation \cite{vitchyr}. For each experiment, 5 random seeds were used per method and domain, with 3 used for the transition policies experiments.

It is recommended that if this algorithm is recreated, the batch size is at least as large as in this paper, to ensure that the convex combination of critics is effective across each batch. It is also recommended that the discount factor \(\gamma\) is not increased above 0.95, to ensure the method doesn't become unstable when estimating too far into the future.
\color{black}

The transition policies baseline was trained using the TRPO\cite{schulman2015trust} algorithm within the implementation provide with its publication \cite{lee2018composing}. The plots in this paper that show the performance of the TP algorithm only represent the training of the transition policies themselves. The primitive policies that learned to solve each subtask were trained separately, and then each run that the TP algorithm underwent used this same set of primitive policies. 

\begin{table}[H]
    \begin{center}
    \caption{Hyperparameters used in Maze Domain}
    \label{tab:Hypers}
    \begin{tabular}{cc}
        \toprule
        \textbf{Hyperparameter} & \textbf{Value} \\
        \midrule
        Gamma (\(\gamma\)) & 0.95 \\
        Replay buffer length & 1e6 \\
        Batch size & 256 \\
        Maximum episode length & 1000 \\
        Soft target update factor & 0.005 \\
        Timesteps per epoch & 5000 \\
        Training loops per epoch & 1000 \\
        \bottomrule
    \end{tabular}
    
    \end{center}
\end{table}

\newpage
\section{Appendix 2} \label{app:proof}
This section proves that using a convex combination of two critic functions can be used for the purposes of training a policy to solve a task.

\subsection{Preliminaries}
Consider a task that has an action space \(A\), a state space \(S\), and a reward signal \(R\). This task can be decomposed into a series of \(N\) subtasks such that each subtask has an action space \(A_n \subseteq A\), a state space \(S_n \subseteq S\), and a reward signal \(R_n \in [0,1]\). Each subtask reward signal is defined by,
\begin{equation}
    R_n(s,a) = 
    \begin{cases}
    \in [0,1],& \text{if } s \in S_n\\
    0,& \text{else if } s \in S_m | m \in [1,n) \\
    1,& \text{else if } s \in S_m | m \in (n,N]
    \end{cases}
\end{equation}
The reward signal \(R\) for the overall task is defined as,
\begin{equation}
    R(s,a) = \sum_{n=1}^N R_n(s,a) 
\end{equation}
which is equivalent to,
\begin{equation}
    R(s,a) = n-1 + R_n(s,a) \quad | \quad s \in S_n, \forall a \in A
\end{equation}
\color{black}
The environment also contains a subtask transition function \(U(s) \in [1,N)\) that denotes the current subtask. This function will only transition from subtask \(n\) to \(n+1\) when \(R_n \geq 1-e\), where \(e\) is some sensitivity metric.

A Q-function \(Q\) approximates the expected discounted sum of future rewards,
\begin{align}
    Q(s_t,a_t) &= R(s_t,a_t) + \gamma Q(s_{t+1},\pi(s_{t+1}))\\
    &=\sum_{i=t}^\infty \gamma^{i-t}R(s_i,a_i), \quad a_i \sim \pi(s_i)
\end{align}
where \(\pi\) is a policy chosen to maximise the Q function and the environment transitions deterministically.

\subsection{Theorem}

\textit{A convex combination of two normalised Q functions can be used in the place of the sum of two Q functions for the purposes of picking an optimal action. The original maximisation objective can be recovered using the correct weighting.}

A standard policy \(\pi\) selects actions that maximise a Q function such that,
\begin{equation}\label{eq:policyMax}
    Q(s,\pi(s)) = \max_{a \in A}Q(s,a),
\end{equation}
This can be decomposed into two Q functions according to Lemma 1.

\subsubsection{Lemma 1}
\textit{A Q function for a task involving multiple sequential subtasks can be be approximated as the sum of two Q functions, each corresponding to the current and subsequent subtasks}

Assume that for each subtask there exists a Q function \(Q_n\). Assume that the cooperative reward signal \(r^{coop}_n\) for the current subtask \(n\) is given by

\begin{equation}
    r^{coop}_n = \eta r_n + (1-\eta)r_{n+1}
\end{equation}
where \(r_n\) is the reward signal from the environment for subtask \(n\), and \(\eta \in [0,1]\) is a weighting parameter.

Using this assumed reward signal, the cooperative Q function \(C_n\) for a subtask \(n\) can be found.

\begin{align}
    C_n^\pi(s,a) &= \mathbb{E}\left[\sum^\infty_{t=0}\gamma^tr^{coop}_n|\pi,s,a\right]\\
    \begin{split}
    &= \eta\mathbb{E}\left[\sum^\infty_{t=0}\gamma^tr_n|\pi,s,a\right] + ...\\
    &\qquad (1-\eta)\mathbb{E}\left[\sum^\infty_{t=0}\gamma^tr_{n+1}|\pi,s,a\right]\\
    \end{split}
    \\[2ex]
    &=\eta Q_n^\pi(s,a)+(1-\eta)Q^\pi_{n+1}(s,a) \blacksquare
\end{align}

This formulation of a cooperative Q function uses the weighted sum of two Q functions for consecutive subtasks. The subsequent subtask will always exhibit a smaller Q value, due to the rewards of the subtask being further away through time. The above formulation for a cooperative Q function uses the weighting variable \(\eta\) to represent both this natural weighting of time discounting of rewards, as well as the weighting that represents balancing between the current and subsequent subtasks. To remove this natural time discounting, a normalised cooperative Q function \(\hat{C}_n\) is introduced.

\begin{equation}
    \hat{C}_n= \eta \hat{Q}_n(s,a) + (1-\eta) \hat{Q}_{n+1}(s,a),
    \label{eq:ConvexCombination}
\end{equation}
where \(\hat{Q}\) is a normalisation such that,

\color{black}
\begin{equation}
    \hat{Q}(s,a) = \frac{Q(s,a)-\displaystyle\min_{a\in A} Q(s,a)}{\displaystyle\max_{a\in A} Q(s,a)-\displaystyle\min_{a\in A} Q(s,a)}
\end{equation}
To simplify this operation, the following functions are defined,
\begin{equation}
    Ran(Q) = \max_{a\in A} Q(s,a)-\min_{a\in A} Q(s,a)
\end{equation}
\begin{equation}
    Min(Q) = \min_{a\in A} Q(s,a)
\end{equation}
Using these it can be shown that using the correct \(\eta\) can lead to the same action.
\begin{align}
    &\max_{a\in A}(\hat{C}_n) =  \max_{a\in A}(\eta \hat{Q}_n + (1-\eta) \hat{Q}_{n+1})\\
    &= \max_{a\in A}\left(\eta \frac{Q_n-Min(Q_n)}{Ran(Q_n)} + (1-\eta) \frac{Q_{n+1}-Min(Q_{n+1})}{Ran(Q_{n+1})}\right)\\
    &=\max_{a\in A}\left( \frac{\eta}{Ran(Q_n)}Q_n +  \frac{(1-\eta)}{Ran(Q_{n+1})}Q_{n+1}\right)
\end{align}
The evaluations \(\max(\hat{C}_n)\) and \(\max(Q)\) produce the same actions when the coefficients of the convex operation are equivalent. Therefore,
\color{black}
\begin{align}
    \frac{\eta}{Ran(Q_n)} &= \frac{(1-\eta)}{Ran(Q_{n+1})}\\
    \eta Ran(Q_{n+1}) &= (1-\eta)Ran(Q_n)\\
    \eta (Ran(Q_{n+1})+Ran(Q_n)) &= Ran(Q_n) \\
    \eta &= \frac{Ran(Q_n)}{(Ran(Q_{n+1})+Ran(Q_n))}
\end{align}
This definition of \(\eta\) is also bounded between 0 and 1. Using this \(\eta\), the original maximisation action can be recovered. \(\blacksquare\)

\end{document}